\documentclass[sn-mathphys,Numbered]{sn-jnl}

\usepackage{graphicx}%
\usepackage{multirow}%
\usepackage{amsmath,amssymb,amsfonts}%
\usepackage{amsthm}%
\usepackage{mathrsfs}%
\usepackage[title]{appendix}%
\usepackage{xcolor}%
\usepackage{textcomp}%
\usepackage{manyfoot}%
\usepackage{booktabs}%
\usepackage{algorithm}%
\usepackage{algorithmicx}%
\usepackage{algpseudocode}%
\usepackage{listings}%
\usepackage{caption}
\usepackage{makecell}
\usepackage{tabularx}

\theoremstyle{thmstyleone}%
\theoremstyle{thmstyletwo}%

\theoremstyle{thmstylethree}%

\begin{document}

\title[]{Edge-Enabled Anomaly Detection and Information Completion for Social Network Knowledge Graphs}


\author[1]{\fnm{Fan} \sur{Lu}}\email{ fanbyxc1014@gmail.com}

\author[1]{\fnm{Huaibin}
\sur{Qin}}\email{hbqin1023@163.com}

\author*[1]{\fnm{Quan} \sur{Qi}}\email{Q.Qi@ieee.org}

\affil[1]{\orgdiv{School
of Information Science and Technology}, \orgname{Shihezi University}, \orgaddress{ \city{Shihezi}, \postcode{832000}, \state{Xinjiang}, \country{China}}}


\abstract{In the rapidly advancing information era, various human behaviors are being precisely recorded in the form of data, including identity information, criminal records, and communication data. Law enforcement agencies can effectively maintain social security and precisely combat criminal activities by analyzing the aforementioned data. In comparison to traditional data analysis methods, deep learning models, relying on the robust computational power in cloud centers, exhibit higher accuracy in extracting data features and inferring data. However, within the architecture of cloud centers, the transmission of data from end devices introduces significant latency, hindering real-time inference of data. Furthermore, low-latency edge computing architectures face limitations in direct deployment due to relatively weak computing and storage capacities of nodes. To address these challenges, a lightweight distributed knowledge graph completion architecture is proposed. Firstly, we introduce a lightweight distributed knowledge graph completion architecture that utilizes knowledge graph embedding for data analysis. Subsequently, to filter out substandard data, a personnel data quality assessment method named PDQA is proposed. Lastly, we present a model pruning algorithm that significantly reduces the model size while maximizing performance, enabling lightweight deployment. In experiments, we compare the effects of 11 advanced models on completing the knowledge graph of public security personnel information. The results indicate that the RotatE model outperforms other models significantly in knowledge graph completion, with the pruned model size reduced by 70\%, and hits@10 reaching 86.97\%.}


\maketitle

\section{Introduction}\label{sec1}
\noindent

Law enforcement agencies have consistently utilized data analysis to aid in daily management and investigative processes, particularly achieving significant advancements in the realm of public safety \cite{van2017beyond}. Properly storing and easily accessing data can enhance individual analytical, judgment, and decision-making capabilities \cite{jie2022risk}. However, events related to collective security still necessitate law enforcement agencies to have a clear understanding of the social relations within the community.

In the early stages, information processing technologies were not sufficiently advanced, resulting in highly fragmented individual information. Law enforcement agencies lacked in-depth analysis of social relations among individuals, leading to the belated perception of many potential correlations only after risks or incidents occurred. This reliance on experienced law enforcement personnel and manual intervention for extensive data matching became prevalent. However, with the rapid development of big data, artificial intelligence technologies, and hardware computing capabilities, delving into and analyzing social relations has become an essential means for proactively predicting and managing collective security events \cite{li2022knowledge}.

Knowledge graphs, as a graphical knowledge representation technology describing entities, relations, and facts, are frequently employed in the analysis of interpersonal relations. They constitute a semi-structured data model composed of a set of entities and the relations among them. The structure of Knowledge graphs is typically built upon ontologies, aiming to present the semantics and associations of information, facilitating machines in better understanding and reasoning about this information. Knowledge graphs can be utilized to represent individuals, organizations, groups, or concepts in social networks, elucidating the relations between them. This aids in the discovery of potential influencers, information propagation paths, and group structures within social networks.

Most artificial intelligence models employed for exploring social relations in public safety heavily rely on traditional cloud-centric architectures. In a cloud-centric framework, all data needs to be uploaded to remote servers for processing, potentially raising concerns about data privacy insecurity. If these servers are subjected to hacking attempts, there is a risk of personal information leakage. Additionally, this approach introduces certain delays and consumes a significant amount of network bandwidth \cite{ranaweera2021survey}. However, edge computing involves processing data and computations close to the data source, resulting in faster data processing and lower latency. In the realm of public safety, edge computing primarily processes data on the devices generating the data, rather than transmitting all data to the cloud. Retaining data locally can significantly mitigate the risk of data leaks and enhance privacy security in the field of public safety.

This study aims to uncover latent information within social relations and present complex individual data networks in a user-friendly visual format. Through edge deployment, real-time processing and data privacy are ensured \cite{yang2022deep}. Leveraging the mined data for anomaly detection and information completion are crucial for law enforcement agencies to achieve efficient social management and identify potential risks. This paper applies a knowledge graph-based social network analysis method within the context of public safety, utilizing an edge architecture. The primary contributions of this paper are summarized as follows:

\begin{itemize}
  \item Through a three-step process involving ontology construction, information extraction, and knowledge fusion, data is transformed into structured knowledge. This transformation enables key information to be accurately and rapidly retrieved.
  \item Leveraging KGE technology, this study aims to infer missing knowledge within existing data, identify erroneous information, and perform anomaly detection and completion on the current dataset.
  \item The optimal model, determined through comparative analysis, was subjected to model pruning using the proposed pruning algorithm in this paper. This led to a 70\% reduction in the model size. After fine-tuning, the HITS@10 metric only experienced a marginal decrease of 2 percentage points.
  \item Integrating edge computing technology, the pruned model is deployed on edge devices. Processing data at the edge ensures real-time capabilities while, to a certain extent, safeguarding data privacy and security.
\end{itemize}

Remaining sections of the paper are outlined as follows. Section 2 provides a detailed exposition of related work. Section 3 introduces the 11 models employed in the study. In Section 4, data processing, ontology construction, and data fusion are conducted, presenting a novel data quality assessment method named PDQA and a model pruning approach. Section 5 conducts comparative experiments on the graph constructed in the previous section using the 11 advanced models, selecting the best-performing model for pruning, deployment on edge devices, and presenting experimental results. Section 6 offers a concluding summary of the paper.

\section{Related Work}
\subsection{Social Network Knowledge Graph}
\noindent

Social network knowledge graphs are employed to describe, represent, and analyze entities and relations within social networks. They utilize graph structures to visualize and structure various entities in social networks along with their interconnections. Presently, numerous scholars both domestically and internationally have delved into the potential value of social networks.

Social network analysis technology, as a scientific method for studying society, has played a significant role in advancing intelligent law enforcement and public safety \cite{xu2022safe}. Initially, the field primarily focused on the analysis of network quantities and visual representation. Bourmpoulias et al. \cite{bourmpoulias2023entity} explored the visualization of knowledge graph associations and personal information. Huang et al. \cite{Constructionofpublic} conducted a comprehensive analysis of public safety events and developed a visualization system to assist law enforcement agencies in understanding cases. Zhu et al. \cite{zhu2023graph} utilized a label propagation algorithm to aid in training edge weights of knowledge graphs, creating a predictive method for law enforcement cases. They improved traditional convolutional neural networks into multi-channel networks to adapt to various case features. He et al. \cite{he2020constructing} proposed a method for de-anonymization and privacy inference based on knowledge graphs, demonstrating its effectiveness on real social network data. Almoqbel et al. \cite{almoqbel2019computational} used neural network models to analyze social networks of terrorist organizations, showcasing strong capabilities in predicting terrorist events.

\subsection{Public Safety System Under Edge Architecture}
\noindent

Current public safety systems are typically built on cloud computing models. However, as the scale expands, they face challenges in real-time data processing. Consequently, public safety systems based on edge computing are gradually becoming a mainstream approach. Edge computing is a distributed computing paradigm that utilizes network technology to shift computing tasks and data storage from the cloud to the network's edge \cite{xu2023cnn}.

The public safety domain leverages various types of edge big data, including government and media datasets, physical spatial data collected by sensors, cyberspace data, and text data containing personal information. Evangelos et al. \cite{maltezos2021public} explored trends, demands, and technologies for public safety in smart cities under the concept of edge computing, proposing an edge computing platform called DECIoT. Ugli et al. \cite{ugli2023cognitive} introduced a novel cognitive video surveillance management framework with a Long Short-Term Memory (LSTM) model. The aim is to reduce standby GPU memory usage through model offloading, saving memory resources. Walczak et al. \cite{walczak2023acceptance} gathered information about individuals with mobility issues using IoT devices to address safety concerns for residents. Zhang et al. \cite{zhang2023internet} proposed an IoT access control scheme based on permissioned blockchain and edge computing, leveraging the advantages of both technologies. This approach involves conducting user identity authentication at the edge, which helps reduce latency and response times, while ensuring security and reliability. Xu et al. \cite{xu2022non} deployed deep learning models to edge devices to meet their computational needs while ensuring low latency. Finding effective optimization methods for deploying public safety models at the edge is of great significance.

Despite the favorable outcomes achieved by existing methods, they have not fully exploited the connections and influences between relations, resulting in some semantic information gaps. Additionally, these methods incur significant computational and storage resource overheads, making deployment on edge devices impractical. Therefore, this paper adopts a knowledge graph-based social network analysis approach applied to the public safety domain. This method integrates KGE technology to infer and discover missing knowledge within existing data, identify erroneous information, and perform error detection and enhancement on existing information to improve its accuracy. Furthermore, the model undergoes pruning and compression to make it lightweight. This significantly reduces the model's demands on computational and storage resources, enabling deployment on edge devices.

\section{Preliminary Knowledge}
\noindent

In this paper, we employed 11 advanced models for comparison to select the most suitable model for personnel information data in the field of public safety. These models include TransE, TransR, AutoSF, DistMult, ComplEx, HolE, KG2E, RotatE, PairRE, RGCN, and TuckER.

The TransE model \cite{bordes2013translating} is based on the distributed vector representation of entities and relations. It continuously adjusts vectors to make the sum of the head entity vector and the relation vector as close as possible to the tail entity vector. The TransR model \cite{lin2015learning}, built upon TransE, projects entities from the entity space to the corresponding relation space. It then establishes connections between projected entities to learn embeddings. The AutoSF model \cite{zhang2020autosf} automatically designs SFs  for different knowledge graphs. It introduces a greedy search algorithm to enhance algorithm speed through filters and predictors.

The DistMult model \cite{yang2015embedding} utilizes a product-form scoring function to measure semantic correlations between entities and relations. The ComplEx model \cite{trouillon2016complex} extends DistMult into complex space to better model asymmetric and reversible relations. The HolE model \cite{nickel2016holographic} uses circular correlation operations instead of the matrix product of relation matrices. The KG2E model \cite{he2015learning} uses a Gaussian distribution instead of vectors for KGE, limiting the covariance matrix of the distribution to a diagonal matrix to measure similarity. The RotatE model \cite{sun2018rotate} defines each relation as a rotation in complex vector space and proposes a self-adversarial negative sampling technique.

The PairRE model \cite{chao2021pairre} uses two vectors for relation representation, projecting head and tail entities into Euclidean space to minimize the distance of the projection vectors. The RGCN model \cite{schlichtkrull2018modeling} mimics Graph Convolutional Network (GCN), defining node representations. It is designed with different classifiers, encoders, and decoders to flexibly handle various types of graph data. The TuckER model \cite{balazevic2019tucker} can decompose a three-way tensor into a matrix obtained by multiplying each dimension of the core tensor by a matrix, fully representing true triples on the entity set E and relation set R.

\section{Proposed Methodology}
\subsection{Data Source and Ontology Construction}
\noindent

This paper is based on information provided by a municipal public security bureau and constructs a personnel relation dataset, including individual profiles, interpersonal relations, and employment details, among other aspects. 

\begin{figure}[h]
  \centering
  \includegraphics[width=13cm]{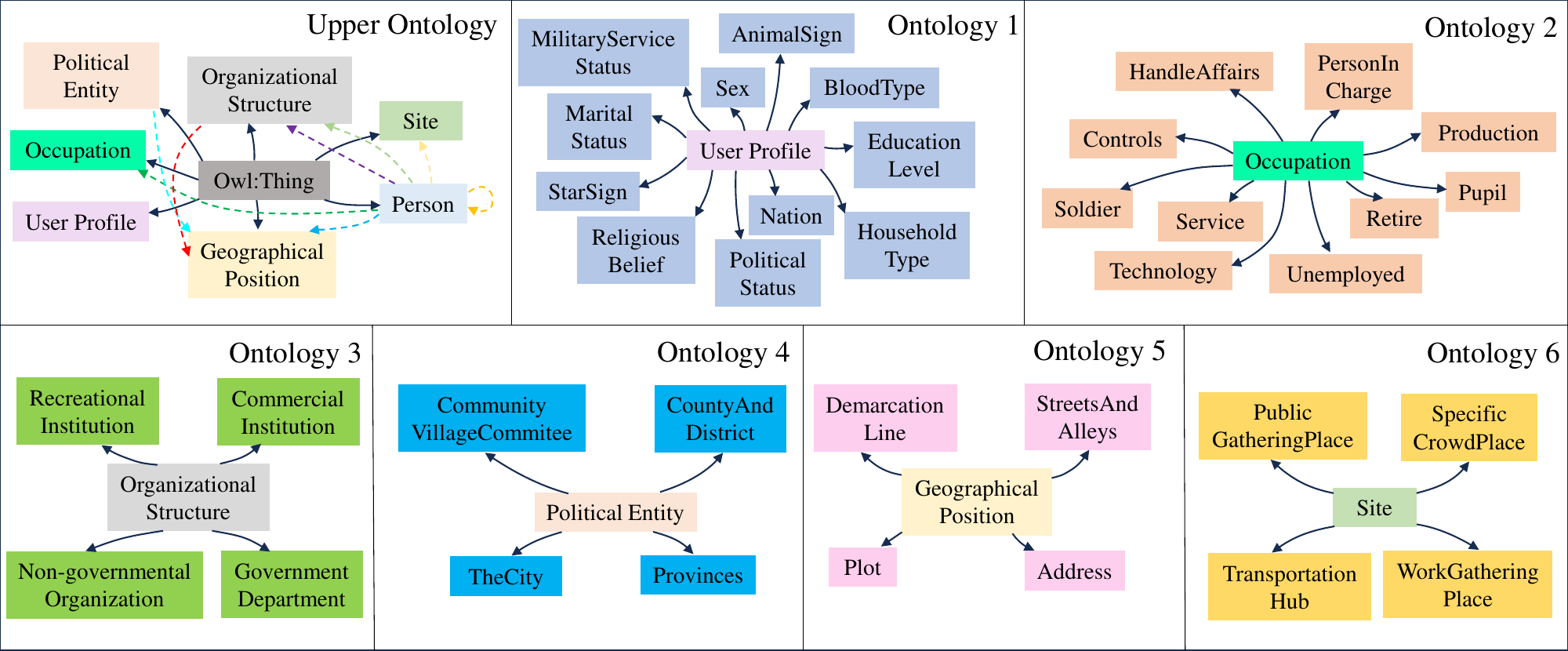}
  \captionsetup{justification=centering} 
  \renewcommand{\figurename}{Fig.}
  \caption{Knowledge graph ontology.}
  \label{Fig:1}
\end{figure}

Firstly, following the classification of data information and business logic, the knowledge graph's ontology is constructed. By considering relevant factual concepts, an abstract model consistent with the business logic is established. In this model, information such as individuals, occupations, and locations is explicitly defined as entities. Fig. 1 illustrates the ontology structure of the knowledge graph construction.

   

\subsection{Information Extraction and Knowledge Fusion}
\noindent

Based on the previously identified entities and relations, we conduct data cleaning on the disorganized dataset. This process includes deduplication, handling missing values, addressing anomalies, and data normalization. We utilize Pandas for dataframe operations, employ regular expressions for text matching and processing, and adopt the PDQA method proposed in this paper to identify anomalies and missing values, subsequently correcting the abnormal data through manual intervention. The data entries subjected to these processes are systematically classified and merged. Ultimately, we identify eleven distinct occupations, ten different education levels, and six marital statuses. Fig. 2 illustrates the distribution of education levels, clearly showing a significant decrease in population with higher education levels. Fig. 3 presents the distribution of marital statuses, indicating that the majority of individuals are either married for the first time or unmarried, showing a similar distribution pattern. Fig. 4 displays the distribution of occupations, with the "Unknown" category having the highest count.

\begin{figure}[h]
  \centering
  \includegraphics[width=11cm]{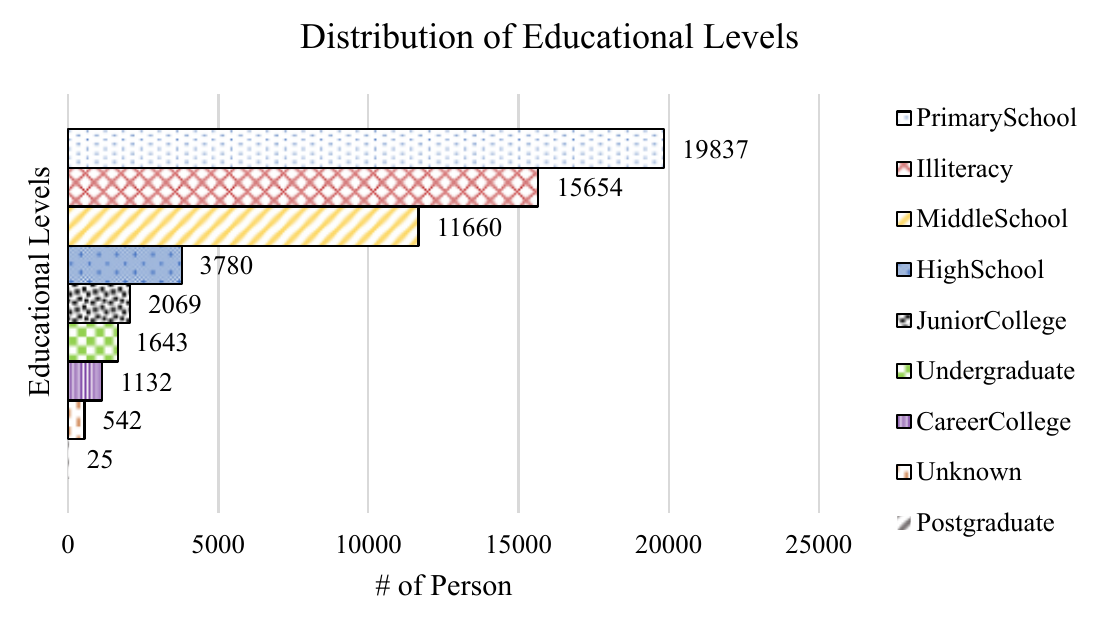}
  \captionsetup{justification=centering} 
  \renewcommand{\figurename}{Fig.}
  \caption{Personnel information statistics (education level).}
  \label{Fig:2}
\end{figure}

\begin{figure}[h]
  \centering
  \includegraphics[width=10cm]{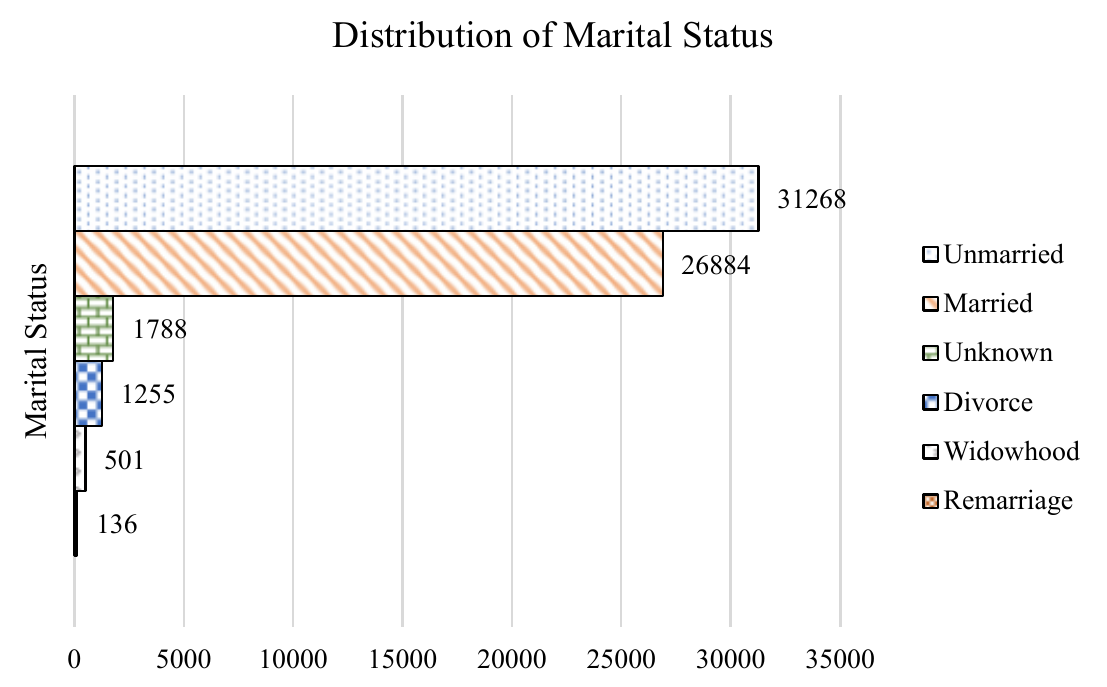}
  \captionsetup{justification=centering} 
  \renewcommand{\figurename}{Fig.}
  \caption{Statistical chart of personnel information (marital status).}
  \label{Fig:3}
\end{figure}

\begin{figure}[h]
  \centering
  \includegraphics[width=12cm]{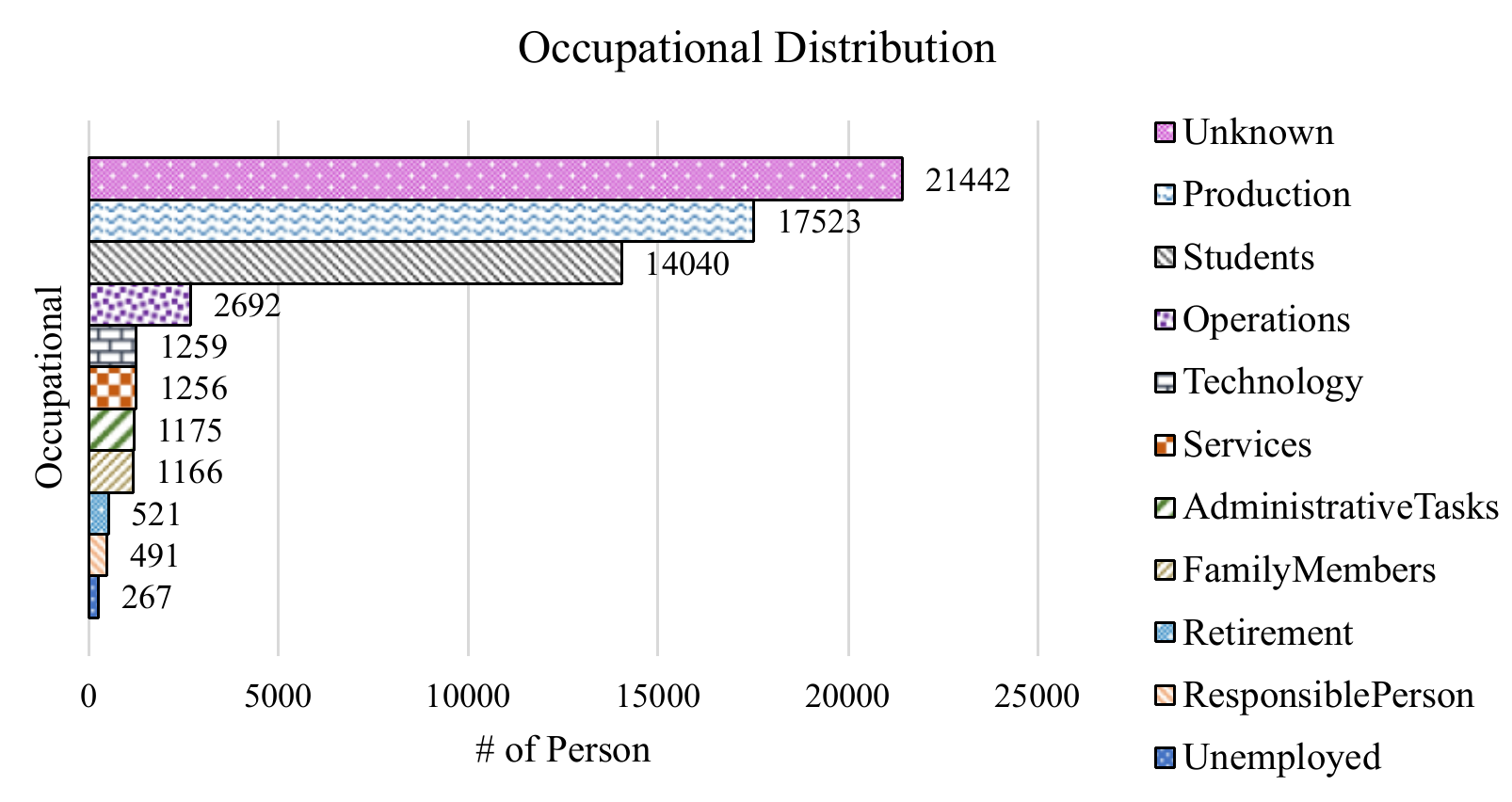}
  \captionsetup{justification=centering} 
  \renewcommand{\figurename}{Fig.}
  \caption{Personal information data (occupational).}
  \label{Fig:4}
\end{figure}

Knowledge fusion involves integrating knowledge from diverse sources and domains into a unified representation. In this study, we conducted knowledge fusion on multi-source data using information such as names, phone numbers, and family relations. This process involved matching entities from different knowledge graphs or data sources to consider them as the same entity, ensuring the uniqueness of entity names for the same concept. The result produced a dataset consisting of 997,956 triples, which will be used in subsequent experiments involving KGE.

\subsection{Knowledge Graph Visualization}
\noindent

By using Py2Neo to input the triples into the Neo4j graph database, data visualization has been achieved, preparing for subsequent inferential computations. Neo4j is a graph database that stores data in a graph structure and offers efficient graph algorithms for data processing. Py2Neo is a library for Python that interfaces with the Neo4j graph database, providing a convenient interface to make working with Neo4j in Python easier. This knowledge graph comprises entities, relations, and attributes. Entities are represented as nodes, with a total count of 61,857, while relations are represented as edges, totaling 140,219. Additionally, there are 87,632 attributes, and their specific types are detailed in Table 1.

\begin{table}[h]
\renewcommand{\tablename}{Table}
\caption{Category of tags and their quantity and type.}
  \label{tab:2}  
  \begin{tabular}{lll} 
    \toprule
    Category  & Quantity & Introduction \\
    \midrule
      Entities (Nodes) & 61857 & Individuals, Occupations, Locations, etc. \\
      Relations (Edges) & 140219 & Familial Relations, Occupational Connections, etc. \\
      Attributes & 87632 & Age, Height, Weight, Blood Type, etc.  \\
     Labels & 83 & Gender, Educational Level, Political Affiliation, Marital Status. \\
   
   \bottomrule
\end{tabular}

\end{table}

\subsection{Evaluation Methods}

\subsubsection{Algorithm Performance Evaluation Method}
\noindent

After the completion of the construction of the personnel information knowledge graph, it is necessary to complete the information through KGE. 
In KGE algorithms, the performance of these algorithms is typically evaluated using the following set of evaluation metrics:

(1) HITS@N: HITS@N refers to the proportion of correct answers ranked within the top K positions in a ranking task \cite{paulheim2017knowledge}. For instance, in a ranking task with 10 samples, if the ranks of the 5 correct answers are 1, 2, 3, 4, and 5, then when N=5, the value of HITS@N is 5/5=1; when N=6, HITS@N is 5/6=0.83. The mathematical expression is as follows: 

\begin{equation}
HITS@N = \frac{1}{n} \sum_{i=1}^{n} [rank_{i} \le N]
\label{eq:1}
\end{equation}

In this context, where $n$ represents the number of samples, and $rank_{i}$ signifies the ranking of the $i-th$ sample, [ ] denotes an indicator function for conditional evaluation (with a value of 1 if the condition is true, and 0 otherwise). A higher HITS@N value indicates better algorithm performance.

(2) AMRI (Adjusted Arithmetic Mean Rank Index): Introduced by Berrendorf, he pointed out that MR and HITS@N increase proportionally with the reduction of the test set size \cite{huang2022cross}. Similarly, MRR and HITS@N also exhibit high homogeneity. Therefore, he proposed a new evaluation metric called AMRI. The mathematical expression is as follows:

\begin{equation}
AMRI = \frac{2 {\textstyle \sum_{i=1}^{n}}(rank_{i}-1) }{ {\textstyle \sum_{i=1}^{n}}(\mid s_{i}\mid ) }
\label{eq:2}
\end{equation}

In this context, where $n$ denotes the number of samples, $rank_{i}$ represents the ranking of the $i-th$ sample, and $S$ signifies the score assigned to the $i-th$ sample.

The AMRI metric ranges from -1 to 1, with values closer to 1 indicating better model performance and values less than 0 suggesting that the model performs worse than random scoring.

(3) Standard Deviation: The standard deviation of rankings \cite{lee2015standard}. The mathematical expression is as follows:

\begin{equation}
\delta =  \sqrt{\frac{1}{n} \sum_{i=1}^{n}(rank_{i}-\mu )^{2} } 
\label{eq:3}
\end{equation}

In this context, where $n$ denotes the number of samples, $rank_{i}$ represents the ranking of the $i-th$ sample, and $\mu$ represents the average ranking of the samples.

(4) Time: This includes both training time and prediction time, measured in seconds. The training and prediction times can reflect the deployment cost and practical value of the model, preparing for the model's application and deployment.

\subsubsection{Data Evaluation Method PDQA}
\noindent

To detect existing anomalous data and reduce the cost of manual screening, this paper introduces a data quality assessment method called PDQA. The algorithm aims to assess the quality of existing data while also verifying the compliance of future input data. The structure of PDQA is depicted in Fig. 5.

\begin{figure}[h]
  \centering
  \includegraphics[width=13cm]{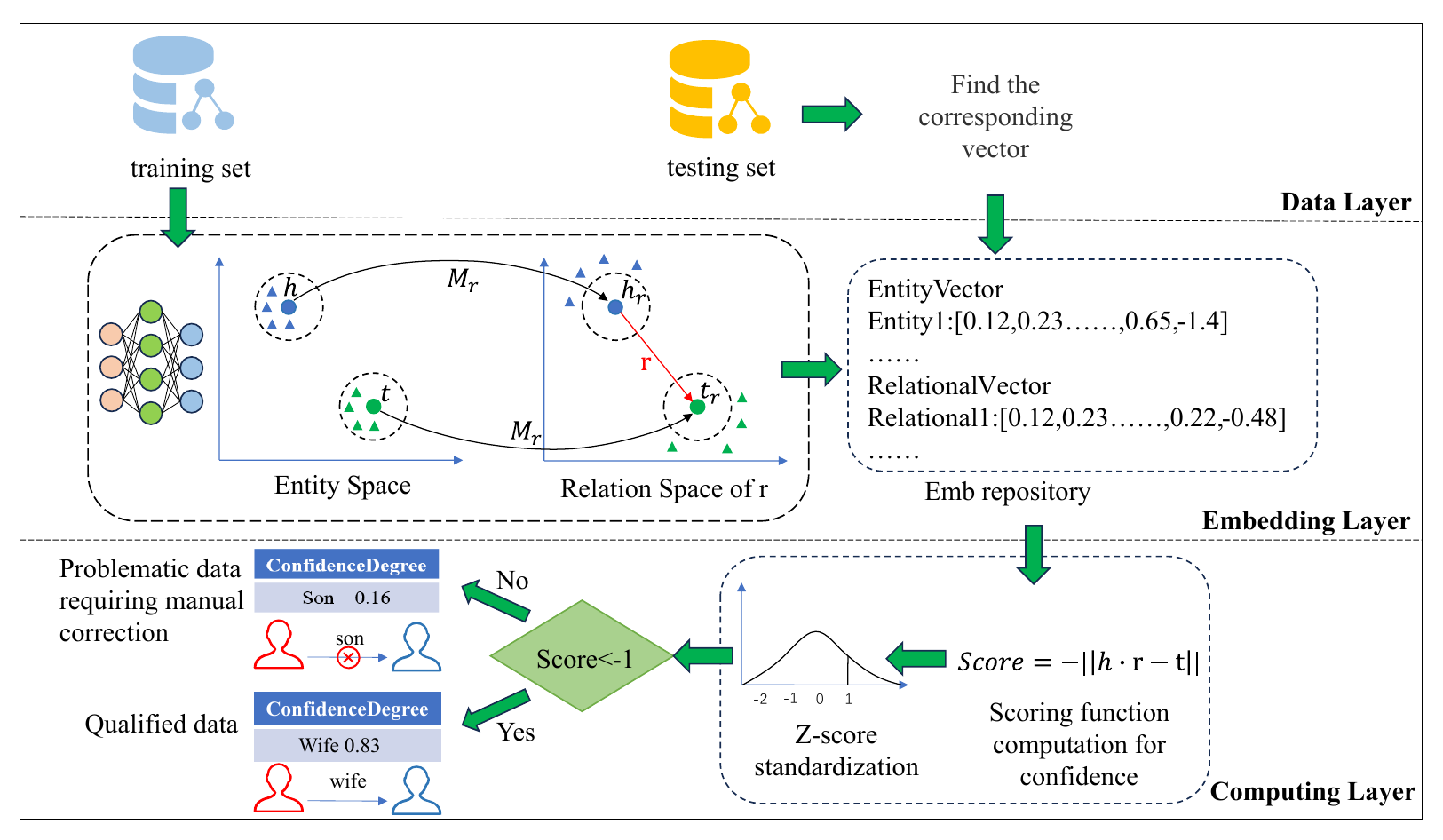}
  \captionsetup{justification=centering} 
  \renewcommand{\figurename}{Fig.}
  \caption{PDQA structure diagram.}
  \label{fig:5}
\end{figure}

In the PDQA algorithm, the training data undergoes training with the KGE model to obtain corresponding embedding vector representations. Subsequently, for the test data and simulated error data, we query the respective vector representations. The model's corresponding scoring function is used to evaluate the confidence of these vectors. Finally, standardization is performed using the $z-score$. Compared to Min-Max normalization, $z-score$ normalization is more robust to outliers. It uses the mean and standard deviation, not influenced by the maximum and minimum values, making it more suitable for handling data with outliers. Compared to Decimal Scaling normalization, $z-score$ normalization better preserves the distribution shape of the data, as it is based on the mean and standard deviation, preserving the original distribution characteristics of the data. In contrast, Decimal Scaling simply divides by a fixed value. Therefore, we choose $z-score$ for standardization here. After standardization, data with $z-score$ values less than -1 are considered problematic and should be reported for further investigation and manual confirmation. The formula for $z-score$ normalization is as follows:

\begin{equation}
z = \frac{x- \mu}{\delta } 
\label{eq:4}
\end{equation}

In this formula, where $x$ represents the original data value, $\mu$ denotes the sample mean, $\delta$ stands for the sample standard deviation, and $z$ signifies the standardized value. The pseudocode for the PDQA evaluation method is as follows:

\begin{algorithm}
\caption{PDQA evaluation algorithm.}\label{algo1}
\begin{algorithmic}[1]
\Require Raw data that needs to be filtered.
\Ensure Problematic data requiring manual screening. 

\State Calculate embeddings for the personnel data training set using the KGE model and store them in the embedding library.
\State Find the corresponding vector of the personnel data test set in the embedding library.
\State Calculate confidence through scoring function.
\State \textbf{initialize} 

$x \Leftarrow $ Raw value data.

$\mu \Leftarrow $ Sample mean.

$\delta \Leftarrow $ Sample standard deviation.

\State $z-score = (x-\mu) / \delta  \Leftarrow $ Normalized values.

\If{$z-score < -1$}\label{algln2}
        \State $PD \Leftarrow $ It is problematic data that requires manual correction.
\Else
        \State Qualified data.
\EndIf      
\State \Return $PD$
\end{algorithmic}
\end{algorithm}

The PDQA algorithm allows for a comprehensive and objective assessment of data quality, enabling effective screening. Compared to traditional manual screening methods, PDQA has several advantages. Firstly, through an automated process, PDQA significantly improves screening efficiency, saving time and human resources. Secondly, the use of standardized evaluation metrics in the PDQA algorithm provides more accurate and reliable data quality assessments, avoiding the influence of subjective factors. Additionally, the PDQA algorithm is capable of efficiently screening large-scale data, enhancing the ability and speed of data processing. Lastly, PDQA can help establish a standardized management system for data quality, strengthening data governance capabilities and thereby elevating the overall level of data management.

\subsection{Model Pruning Algorithm}
\noindent

In the model pruning algorithm proposed in this paper, the process begins by loading the pre-trained model. Subsequently, the sensitivity of each parameter is computed using the validation set, where sensitivity is measured by the gradient of the parameters \cite{augasta2013pruning}. A larger absolute gradient indicates a greater impact of the parameter on the model's performance. A sensitivity threshold is set based on the percentile of the gradient metric, and parameters with sensitivities below this threshold are set to zero. Finally, the pruned model undergoes fine-tuning to enhance its performance after pruning. The pseudocode for the model pruning algorithm is as follows:

\begin{algorithm}
\caption{Model pruning algorithm.}\label{algo2}
\begin{algorithmic}[1]
\Require Trained model file and validation dataset.
\Ensure Pruned model. 

\State Load the trained model.
\State Evaluate the model on the validation set and obtain the sensitivity of each parameter.
\State Set the threshold of sensitivity.
\For {layer} {
\If{This layer contains trainable weights.}
        \State Get the weights of this layer.
        \State Create a mask based on the threshold value.
        \State Pruning with masks.
        \State Update model weights.
\EndIf     
}
\EndFor
\State $Prmodel$ $\Leftarrow$ Save the pruned model file.
\State \Return $Prmodel$
\end{algorithmic}
\end{algorithm}

The core idea of the model pruning algorithm proposed in this paper is to leverage sensitivity information of parameters. By removing parameters with relatively minor impact on the model, the complexity of the model is reduced while attempting to preserve its predictive performance. This approach, compared to straightforwardly truncating parameter values, gives more consideration to the contribution of parameters to the overall performance of the model. As a result, it can perform pruning in a more intelligent manner to some extent.

\section{Experiment and Analysis}
\subsection{Experiment Overview}
\noindent

In this paper, experiments are conducted using the constructed personnel information knowledge graph. This graph includes a subset of social network relations, basic personal information, employment details, and other fundamental features. The goal is to predict and complement missing information. Additionally, the method proposed in this paper also detects relations that should not exist, specifically those with excessively low confidence, as part of the data anomaly detection process. The schematic diagram of the information completion and detection process is illustrated in Fig. 6.

\begin{figure}[h]
  \centering
  \includegraphics[width=13cm]{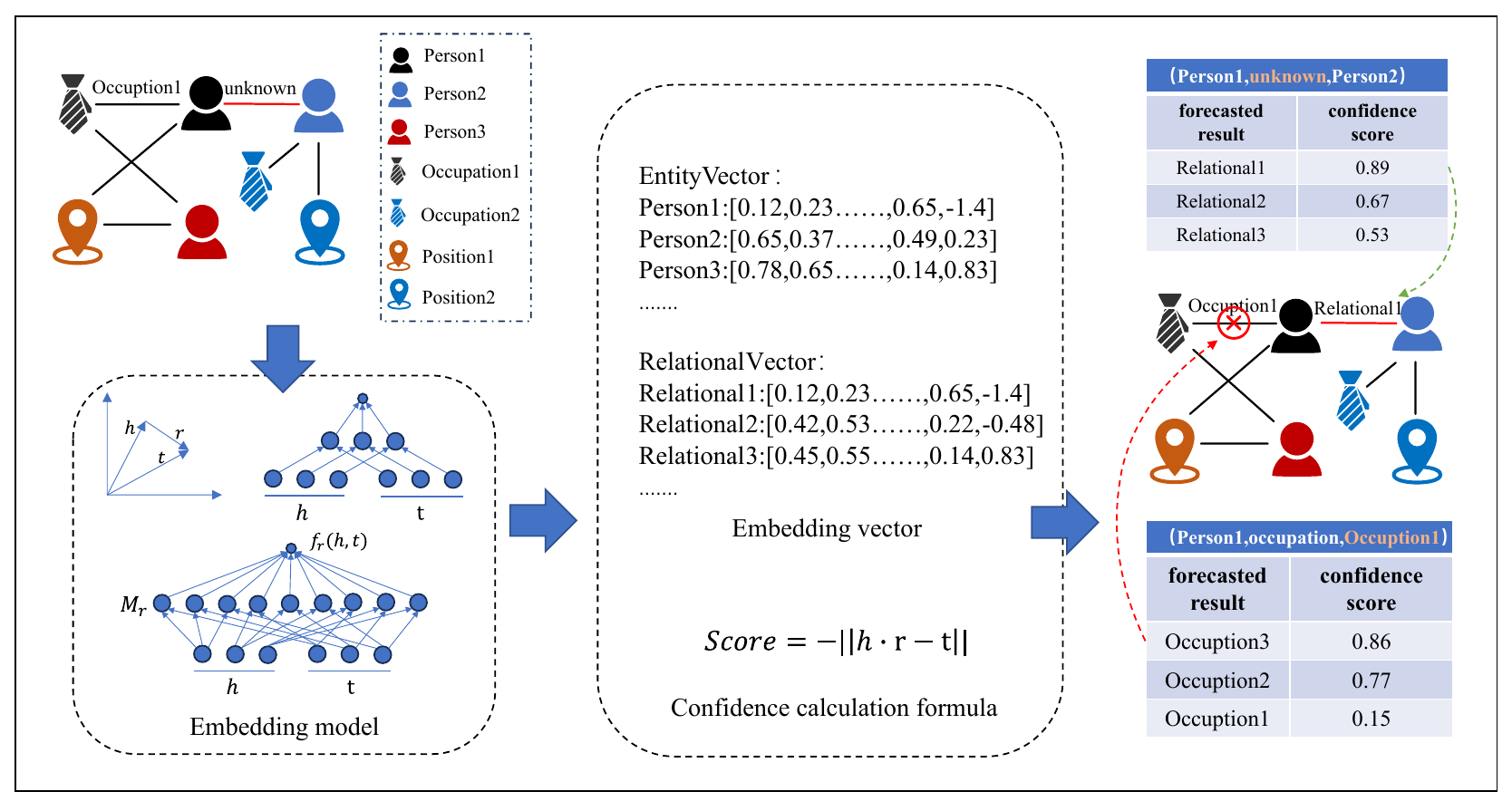}
  \captionsetup{justification=centering} 
  \renewcommand{\figurename}{Fig.}
  \caption{Information completion and anomaly detection schematic diagram.}
  \label{Fig:6}
\end{figure}

In Fig. 6, by embedding entities in the personnel information knowledge graph to obtain their vector representations, we subsequently trained the model on a large number of triple relation vectors. The resulting model can be used to infer the confidence level of relations between two entities. A high confidence level indicates a higher probability of the existence of a relation between these two entities.

\subsection{Experimental Environment and Configuration}
\noindent

This paper implemented the model using the DGL-KE toolkit based on the PyTorch framework \cite{zheng2020dgl}. The toolkit supports both GPU and CPU parallel computation and facilitates training on large-scale graphs through partitioning, making it suitable for subsequent deployment and use. In the experiments, the input to the network consisted of triplets representing social networks, basic information, and job-related details. The dataset was split into training, validation, and test sets in a ratio of 8:1:1. The training set comprised 798,364 triplets, while the validation and test sets each contained 99,796 triplets. The training set was used for model training, the validation set for selecting the best model hyperparameters, and the test set for evaluating model performance.

This paper compares and analyzes eleven models mentioned in Preliminary Knowledge: TransE, TransR, AutoSF, ComplEx, DistMult, HolE, KG2E, PairRE, RGCN, RotatE and TuckER. During the training process, this paper conducts numerous parameter tuning experiments and ultimately determines the following parameters: the dimensions of the embedding vectors for entities and relations are set to 400, the number of negative samples is 50, the learning rate is set to 0.001, and the optimizer used is Adam. Additionally, an L2 regularization coefficient of 0.01 was applied. The training was conducted for 100 epochs, and early stopping was employed to prevent overfitting. Each training batch consisted of 5000 samples, and this training was performed on the aforementioned knowledge graph of interpersonal relations. For the evaluation phase, we assessed the model performance by calculating the rank of each test triplet, where a lower rank indicates better predictive performance. Four categories of metrics were utilized to evaluate the models: Time, AMRI, Standard Deviation, and HITS@N, with N values set at 1, 5, and 10.

The edge device used for deployment is the Jetson Nano. The model is deployed locally, utilizing edge computing to process and analyze data near the collection point. This approach eliminates the need to transmit sensitive data to remote servers, reducing the risk of interception or leakage during the data transfer process. The configuration of Jetson Nano is is presented in Table 2.

\begin{table}[h]
\renewcommand{\tablename}{Table}
\caption{The configuration of Jetson Nano.}
  \label{tab:2}
  \begin{tabular}{ccccc} 
    \toprule
    \makecell{ Development \\ Environment }  & CPU & GPU & RAM \\
    \midrule
    
      \makecell{Python=3.7 \\ C++,CMake} & \makecell{64-bit quad-core \\ ARM A57} &  \makecell{128-core \\ NVIDIA Maxwell } & \makecell{4GB 64-bit \\ LPDDR4 25.6GB/s} \\

   \bottomrule
\end{tabular}

\end{table}

Before being stored, personal identity information undergoes anonymization and de-identification processes. This involves replacing key information with numerical IDs to ensure privacy protection while still enabling effective analysis. Edge devices retain local copies of the data, reducing the need for frequent data requests to central servers and thereby lowering the risk of data exposure \cite{xu2022computation}.

\subsection{Model Comparison Results}
\noindent

The post-training results are presented in the Table 3. Analyzing the experimental results yields the following observations:
\begin{table}[h]
\renewcommand{\tablename}{Table}
\caption{Model performance comparison results.}
  \label{tab:3}
  \begin{tabular}{cccccccc} 
    \toprule
    Model  & \makecell{ Training\\ Time } & \makecell{ Evaluation\\ Time } & AMRI & HITS@1 & HITS@5 & HITS@10 & \makecell{ Standard\\ Deviation } \\
    \midrule
    
    \textbf{}
    
    AutoSF  & 7401.46 & 886.92 & 97.07 & 4.84 & 12.15 & 16.34 & 4268.85 \\
     
    ComplEX  & 3781.66 & 272.82 & 97.22 & 17.41 & 24.94 & 27.91 & 3916.36 \\

    DistMult  & \textbf{1326.91} & 185.95 & 98.97 & \textbf{39.35} & \textbf{64.97} & \textbf{73.92} & 3837.18 \\

    HolE  & 4618.70 & 454.28 & 97.57 & 9.34 & 24.98 & 37.22 & 4937.57 \\

    KG2E  & 2443.01 & 658.22 & 94.33 & 14.29 & 46.27 & 55.50 & 8198.36 \\

    PairRE  & 2431.77 & 239.80 & 99.60 & 23.32 & 54.46 & 65.96 & 1062.17 \\

    RGCN  & 274649.04 &  \textbf{84.14} & \textbf{99.64} & 13.29 & 40.35 & 53.31 & \textbf{895.32} \\

    RotatE  & 6094.38 & 245.88 & \textbf{99.80} & \textbf{40.20} & \textbf{71.83} & \textbf{80.68} & 1271.88 \\

    TransE  & \textbf{1265.07} & \textbf{141.38} & 99.60 & 0.01 & 56.33 & 66.78 & 1272.58 \\

    TransR  & 3928.77 & 264.77 & 97.84 & 9.0989 & 30.60 & 41.10 & 2704.65 \\

    TuckER  & 16731.63 & 778.16 & 82.69 & 0.1144 & 1.170 & 2.160 & 11966.31 \\

   \bottomrule
\end{tabular}

\end{table}

(1) Regarding training and evaluation times, TransE exhibits the best performance, followed by DistMult, while RGCN and AutoSF require the most time.

(2) Concerning the AMRI metric, RotatE and RGCN perform relatively well, with TuckER having the lowest performance at 82.69\%, while other models achieve over 90\%.

(3) In terms of the Hits@1 metric, TransE and DistMult show subpar performance, while HolE performs well, and RotatE stands out as the top performer.

(4) As for Hits@5 and Hits@10 metrics, DistMult and RotatE demonstrate the best performance, while other models exhibit more moderate results.

(5) PairRE doesn't achieve the best performance among all models, but it exhibits a smaller standard deviation, indicating its stable results.

In summary, the RotatE algorithm outperforms other models on these three metrics, but its training and evaluation times are relatively long, requiring a trade-off between performance and time. DistMult and RGCN also demonstrate good results and can be considered as alternative solutions. Additionally, PairRE exhibits good stability. To balance model performance with lower training costs, we chose the RotatE model as the embedding vector generation method. Using a randomized tuning approach, optimal parameters were obtained after 100 iterations on this dataset. We selected an embedding dimension of 976, employed the Adam optimizer with a learning rate of 0.001, and set the batch size to 4096. During the negative sampling phase, we used the Basic negative sampler, generating 8 negative samples for each positive example. The detailed results of parameter tuning can be found in Table 4.

\begin{table}[h]
\renewcommand{\tablename}{Table}
\caption{RotatE tuning effect.}
  \label{tab:4}
  \begin{tabular}{cccccccc} 
    \toprule
    Model  & \makecell{ Training\\ Time } & \makecell{ Evaluation\\ Time } & AMRI & HITS@1 & HITS@5 & HITS@10 & \makecell{ Standard\\ Deviation } \\
    \midrule
     \makecell{ Basic\\ Parameters }  & 6094.38 & 245.88 & 99.80 & 40.20 & 71.83 & 80.68 & 1271.88 \\
      \makecell{ After\\ Tuning }  & \textbf{3594.57} & 550.18 & \textbf{99.87} & \textbf{59.35} & \textbf{83.83} & \textbf{88.87} & \textbf{1148.50} \\

   \bottomrule
\end{tabular}

\end{table}

The experimental results indicate that our proposed approach, combining social network data and utilizing the RotatE model for data completion and detection, is practically feasible.

\subsection{Model Pruning}
\noindent

Utilizing the model pruning algorithm to reduce the model parameters for lightweight processing. During the pruning process, a threshold is calculated based on the pruning rate, and weights with absolute values smaller than this threshold are set to zero. The pruning rate is determined through an experimental process involving parameter tuning by enumeration. Subsequently, the pruned sparse model undergoes fine-tuning. The experimental results are shown in Table 5.

    


\begin{table}[h]
\renewcommand{\tablename}{Table}
\caption{\protect\parbox{\textwidth}{The model performance with different pruning ratios.}}
  \label{tab:5}

  \begin{tabular}{ccccc} 
    \toprule
    \makecell{ Pruning\\ ratio }   & Parameters/M & \makecell{ Model storage\\ volume/MB } & FLOPs/G &  HITS@10 \\
    \midrule
    
    0 & 72.42 & 231.64 & 10.13 & 88.87  \\    
     30\% & 47.63 & 136.37 & 7.24 & 70.34   \\
     50\% & 20.12 & 100.23 & 4.15 & 60.19   \\
     55\% & 16.73 & 83.42 & 3.86 & 50.85  \\
     60\% & 11.57 & 75.98 & 3.21 & 45.34   \\
     67\% & 10.23 & 70.21 & 2.98 & 43.13   \\

   \bottomrule
\end{tabular}

\end{table}

In the initial stage of pruning, the model contains a large number of redundant parameters \cite{peng2021accelerating}. After pruning, many redundant parameters are removed, significantly reducing computational complexity and model storage size. As the pruning ratio increases, more redundant parameters are gradually removed, but the reduction in the number of parameters and model size weakens. When the pruning ratio reaches 67\% and the storage volume decreases by 70\%, the accuracy of the pruned model sharply decreases. This drop is much more significant than the previous decrease, indicating that the pruning ratio is approaching its maximum limit. Further pruning may lead to a significant reduction in model accuracy, and even with subsequent fine-tuning, it may be challenging to fully recover the model performance.

\subsection{Model Fine-Tuning Results}
\noindent

To compensate for the loss in model performance, it is necessary to fine-tune the pruned model \cite{hoefler2021sparsity}. During the fine-tuning process, parameter settings are typically consistent with regular training, with the only exception being the epoch number set to 300 to thoroughly train the model and restore its accuracy. Given the representative nature of the HITS series metrics, especially HITS@10, which more accurately reflects relevant effects, HITS@10 is chosen as the metric for evaluating the model's performance. Table 6 presents a comparison of the model performance before and after fine-tuning.

\begin{table}[h]
\captionsetup{justification=centering}
\renewcommand{\tablename}{Table}
\caption{Performance comparison before and after model fine-tuning.}
  \label{tab:6}
  \begin{tabular}{ccc} 
    \toprule
    \makecell{ Pruning\\ ratio }  & \makecell{ Pruning-HITS@10 } & \makecell{Fine-tuning-HITS@10 } \\
    \midrule
    
    0 & - &  88.87   \\    
     30\% & 70.34 & 87.94    \\
     50\% & 60.19 & 87.44    \\
     55\% & 50.85 & 87.12   \\
     60\% & 45.34 & 87.01    \\
     67\% & 43.13 & 86.97    \\

   \bottomrule
\end{tabular}

\end{table}

From Table 6, it can be observed that when the pruning ratio is relatively low, the model usually can recover to its original accuracy after fine-tuning. However, at this stage, the model still retains a relatively large number of parameters and FLOPs. As the pruning ratio enters the higher range, accuracy begins to decline due to significant information loss in the model.

Based on the above analysis, this paper chooses to deploy the model with a pruning ratio of 67\%, which reduces the storage size by 70\%, to the edge device. At this point, the model's parameters, FLOPs, and storage size are reduced to 10.23M, 2.98G, and 70.21MB, respectively. The Hits@10 reaches 86.97, close to the training accuracy of the unpruned model. This indicates the effectiveness of the model pruning algorithm. While maintaining model accuracy, it significantly reduces the computational load, the number of parameters, and the model storage size \cite{agarwal2022eight}. The pruned model is more streamlined, consuming fewer hardware resources during runtime, making it well-suited for deployment on resource-constrained embedded devices for information completion tasks \cite{xu2022reputation}.

\section{Conclusion}
\noindent

This paper addresses the challenges in handling personnel data in the field of public safety and proposes recommendations for integrating a social network knowledge graph. Building upon existing domestic and international research on interpersonal relations, the paper creates a knowledge graph dataset of personnel relation information. Utilizing this dataset to predict residents' occupational fields and job roles, the results indicate that the personnel relation knowledge graph can effectively assist law enforcement agencies in anomaly detection and data completion. The RotatE model performs well in this task and is further enhanced through model pruning and deployment on the edge device Jetson Nano, improving the timeliness of model processing and data privacy protection. Future research directions may involve integrating relevant datasets and more comprehensive semantic information to enhance prediction accuracy. Exploring hidden relations in social networks through knowledge graphs remains a promising area of investigation.

\section*{Declarations}
\noindent
\textbf{Data Availability Statements}
Due to the confidentiality of this research project, participants of this study did not agree for their data to be shared publicly, so supporting data is not available.

\newpage

\bibliography{ref}

\end{document}